\documentclass{llncs}

\vbadness=\maxdimen

\usepackage{amsmath}
\usepackage{tikz}
\usetikzlibrary{matrix}
\usetikzlibrary{fit}
\newcommand\tm[2][]{\tikz[overlay,remember picture,baseline=(#1.base),inner sep=0pt]\node(#1){$#2$};}
\usepackage{mathrsfs}
\usepackage{svg}
\usepackage{amsfonts}

\usepackage{biblatex}
\bibliography{ref.bib}

\begin{document}
\title{Using special attention improves change point detection}
\titlerunning{COD} 

\author{Anna Dmitrienko\inst{1,2, 3, 4}, Evgenia Romanenkova\inst{2}, Alexey Zaytsev\inst{2}\\
\email{dmitrienko.ae@phystech.edu} }
\authorrunning{Anna Dmitrienko et al.} 
\institute {Institute for Information and Transmission Problems, Moscow\\
\and
Skolkovo Institute of Science and Technology, Moscow
\and
Moscow Institute of  Physics and Technology, Moscow \and 
Institution of Russian Academy of Sciences Dorodnicyn Computing Centre, Moscow}

\maketitle              

\begin{abstract}
The change point is a moment of an abrupt alteration in the data distribution.   Current methods for change point detection are based on recurrent neural methods suitable for sequential data. However, recent works show that transformers based on attention mechanisms perform better than standard recurrent models for many tasks. The most benefit is noticeable in the case of longer sequences. In this paper, we investigate different attentions for the change point detection task and proposed specific form of attention related to the task at hand. We show that using a special form of attention outperforms state-of-the-art results.

\keywords{change point detection, transformer, attention, sequential data}
\end{abstract}

\section{Introduction}

The goal of Change Point Detection (CPD) is to find the moment of data distribution shift. Such tasks appear in different areas, from monitoring systems to video analysis \cite{aminikhanghahi2017survey} to oil$\&$gas~\cite{romanenkova2019real}.
One of the recent methods for CPD \cite{romanenkova2021principled} proves that a recurrent neural network model constructs meaningful representations and solve a problem better than non-principled approaches. 
The state-of-the-art model for sequential data is Transformer~\cite{vaswani2017attention}.  
The main benefit of Transformer's is their ability to
work with long-range dependencies via attention mechanism. Attention matrix specifies at what exact points we should look~\cite{vaswani2017attention}. 
A modification of attention matrix allows faster and more efficient processing of longer sequences~\cite{tay2020long,tay2020efficient}.

We propose specific attention mechanisms that allow efficient work for the change point detection problem.
By considering different autoregression and non-autoregressive attention matrices, we highlight important properties of both the problem and the model based on transformers.
The proposed models work faster and directly incorporate the peculiarities of the problem at hand.
They outperform existing results for the considered change point detection problem in semi-structured sequences of sequential data.

\section{Related work}

Attention mechanism is an important idea in deep learning for sequential data \cite{vaswani2017attention}. It allows usage of the whole sequence and doesn't a model to have a long memory. Instead, at each layer, the model looks at the whole sequence. 

The transformer architecture based on attention shows impressive results in many problems related to the processing of sequential data and in particular NLP~\cite{fursov2021differentiable}, \cite{reis2021transformers} and computer vision~\cite{khan2021transformers}.

However, direct application of this mechanism can be prohibitively expensive: 
the computational complexity of the vanilla attention is $O(s^2)$, where $s$ is the sequence length.
A quest for more computationally effective attention leaded to numerous ideas explored in reviews~\cite{tay2020long,tay2020efficient}.
One of the key ideas in this area is to use so-called sparse attention: instead of the whole attention matrix, we drop parts of it.
This mechanism also allows highlighting a specific part of sequences.

One of the problem statements for sequential data is the change point: we want to detect a change of a distribution in a sequence as fast as possible~\cite{shiryaev2017stochastic}.
Accurate solutions for such problems are vital in different areas, including software maintenance~\cite{artemov2016detecting} and oil$\&$gas industry~\cite{romanenkova2019real}.
Simple statistics-based approaches are sufficient in many cases~\cite{van2020evaluation}.
However, semi-structured data require deep model to provide reasonable quality of solutions ranging from rather simple neural network~\cite{hushchyn2020online} to more complex workflows~\cite{romanenkova2021principled,kail2021recurrent} and problem statements~\cite{sultani2018real}.

For complex multi-step processing of sequences of videos, we still need efficient models that capture the essential properties of data and can detect the change point detection: both in terms of model training and evaluation time and in detection delay in a sequence.
It seems natural to apply the efficient transformers paradigm to such data: 
the attention that pays attention to the recent past is expected to meet these efficiency and quality requirements.

\section{Problem statement}

The solution of this problem with the neural network was investigated in~\cite{romanenkova2021principled}. Its working principle is as follows.

Let a dataset be a set of sequences $D=\left\{\left(X_{1}, \theta_{1}\right), \ldots,\left(X_{N}, \theta_{N}\right)\right\}$, where each sequence $X_{i}$ has the length $T$ and corresponding change point $\theta_i$ is in $[1, \ldots, T]$, if we have a change point in a sequence and $\theta_i=(T+1)$ otherwise.

Let the random process $X^{1: T}=\left\{\mathbf{x}_{1}, \ldots, \mathbf{x}_{T}\right\}$ of length $T$ be given, where $\mathbf{x}_{i} \in \mathbb{R}^{d}$ is an observation at time $i$. The problem is the quickest detection of the true change moment $\theta$ as possible.

Let $p_{t i}$ be predicted probabilities of the change point at a specified time moment $t$ for the $i$-th sequence.

Let $\mathscr{L}_{\text {delay }}$ and $\mathscr{L}_{\text {FA }}$ be defined as follows:

\begin{align}
\mathscr{L}_{delay}(f_\mathbf{w}, D) = \\ \frac{1}{N}\sum_{i=1}^{N}\left(\sum_{t=\theta_{i}}^{\tilde{T}}{(t-\theta_{i})p_{ti}\prod_{k=\theta_{i}}^{t-1}(1 - p_{ki}) +  (\tilde{T}+1-\theta_{i})\prod_{k=\theta_{i}}^{\tilde{T}}(1-p_{ki})} \right) \nonumber
\end{align}\label{eq1}

\begin{equation}\label{eq2}
\mathscr{L}_{FA}(f_\mathbf{w}, D) = - \frac{1}{N}\sum_{i = 1}^N \left( \sum_{t = 0}^{\tilde{T}_ i} t p_{ti} \prod_{k = 0}^{t - 1} (1 - p_{ki}) - (\tilde{T}_i + 1) \prod_{k = 0}^{\tilde{T_i}} (1 - p_{ki}) \right),
\end{equation}
where $\tilde{T}$ is a hyperparameter that restricts the length of the 
considered part of the sequence for delay loss and $\tilde{T} = \min{(\theta_i, T)}$. 

According to \cite{romanenkova2021principled} the  equation (\ref{eq1}) is the lower bound for the expected value of detection delay and
the equation (\ref{eq2}) is lower bound for the expected time to false alarm.

As it was shown in paper above, there is a principled differentiable loss function for solving the CPD problem via neural network $f_\mathbf{w}$:

\begin{equation}
\mathscr{L}\left(f_{\mathbf{w}}, D\right)=\mathscr{L}_{\text {delay }}\left(f_{\mathbf{w}}, D\right)+c \mathscr{L}_{F A}\left(f_{\mathbf{w}}, D\right),
\end{equation}

\section{Methods}
As there is a principled differentiable loss function for solving the CPD problem, we can teach any model with this loss function.
We want to expand the range of these models, and we start with a model such as a transformer. Also, we apply different masks for the source sequence. 
To compare with ~\cite{romanenkova2021principled} we use the RNN model as a
baseline.
Intuitively, it seems that if we have only one point of change, we do not need to look at the entire sequence of data, but we can only consider some part of it. For this, we will use different masks for the input sequence.
\textit{The mask} is a boolean matrix of elements equal to the number of elements in the input sequence. In the place where \textit{True or 1 stands, the value is not considered}, i.e. the mask covers it.
We have chosen several types of masks:

\begin{enumerate}
    \item lower triangular mask ~--- consider only past points, that implies an online working  mode. The mask has the following form:

      \begin{equation}
        \vcenter{\hbox{\tikz[]{
        \matrix[matrix of math nodes, ampersand replacement=\&,
        left delimiter={[}, right delimiter={]}, row sep=0pt, 
        nodes={inner sep=2pt}] (M) {
        \tm[a]{0} \& ~ 1 \& 1 \& \dots \& 1 \\
        0 \& ~0 \& ~1 \& \dots \& 1 \\
        0 \& ~0 \& ~0 \&  \dots \& 1 \\
        \vdots \&~\vdots \&~\vdots \&~\ddots \&\vdots \\
        \tm[b]{0} \&~ 0 \& ~0 \&  \dots \& \tm[c]{0} \\
        }; 
         }}} 
        \end{equation}  
       
        \begin{tikzpicture}[overlay, remember picture]
          \node(x)[fit=(a) (b),inner sep=0pt]{};
          \node(y)[fit=(b) (c),inner sep=0pt]{};
          \filldraw[rounded corners,opacity=.1,orange](x.north west)--(x.south west)--(y.south east)--(y.north east)--(x.north east)--cycle;
        \end{tikzpicture}
    \item $n$-diagonal mask ~--- n elements on the diagonal, so we consider  $\lfloor \frac{n}{2} \rfloor$ right and left neighboring elements. It is some implementation of the "window" method, see more \cite{article}. The matrix for this mask is
    \begin{equation}
        \vcenter{\hbox{\tikz[]{
        \matrix[matrix of math nodes, ampersand replacement=\&,
        left delimiter={[}, right delimiter={]}, row sep=0pt, 
        nodes={inner sep=2pt}] (M) {
            0\dots 0 \&  \&  \&  \&  \&   \\ 
            \& 0\dots 0 \&  \&  \&  \&   \\ 
            \& \& 0\dots 0 \&  \&  \& \\ 
            \&  \&  \& \dots  \&   \& \\ 
            \&  \&  \&  \& 0\dots0  \\
        }; 
        \fill[orange, opacity=0.1, rounded corners]
          (M-1-1.north west) -- (M-1-1.north east) -- (M-5-5.north east) 
        --(M-5-5.south east) -- (M-5-5.south west) -- (M-1-1.south west)
        --cycle;
        }}} 
        \end{equation}

    \item $3$-diagonal mask plus lower triangular with a side of $n$ elements ~--- points from the beginning of the dataset are added to the previous point. It seems that the model based on the initial elements will give the best result, the matrix for this mask is

    \begin{equation}
        \vcenter{\hbox{\tikz[]{
        \matrix[matrix of math nodes, ampersand replacement=\&,
        left delimiter={[}, right delimiter={]}, row sep=0pt, 
        nodes={inner sep=2pt}] (M) {
            0 ~0~ 0 \&  \&  \&  \&  \&   \\ 
            \& 0 ~0~ 0 \&  \&  \&  \&   \\ 
            \& \& 0 ~0~0 \&  \&  \& \\ 
            \&  \&  \& \dots  \&   \& \\ 
            \&  \&  \&  \& 0 ~0~ 0  \\
        }; 
        \fill[orange, opacity=0.1, rounded corners]
          (M-1-1.north west) -- (M-1-1.north east) -- (M-5-5.north east) 
        --(M-5-5.south east) -- (M-5-5.south west) -- (M-1-1.south west)
        --cycle;
        }}}  + \vcenter{\hbox{\tikz[]{
        \matrix[matrix of math nodes, ampersand replacement=\&,
        left delimiter={[}, right delimiter={]}, row sep=0pt, 
        nodes={inner sep=2pt}] (M) {
            1\&  \dots\&  \&  \&  \&   \\ 
            1\& 1 \& \dots \&  \&  \&   \\ 
            0 \& 1 \& 1 \& \dots \&  \& \\ 
            ~\dots \& 0 \& 1 \& 1  \& \dots  \& \\ 
            0\& \dots \& 0\& 1 \& 1 \\
        }; 
        \fill[orange, opacity=0.1, rounded corners]
          (M-3-1.north west) -- (M-3-1.north east) -- (M-5-3.north east) 
        --(M-5-3.south east) -- (M-5-1.south west)  -- (M-5-1.south west)
        --cycle;
        }}} 
        \end{equation}  
 
    \item 1-diagonal mask plus lower  triangular with side of n elements ~--- this is almost the same as the previous point, but now we do not want to look at the adjacent points to the current one
\end{enumerate}

We investigate how the attention mechanisms can improve change point detection problems compared to an RNN model. 

\section{Results}
\subsection{Data}

As the dataset to evaluate our approach, we use sequences of handwritten digits based on generative models trained on MNIST similar to~\cite{romanenkova2021principled}.
To generate this dataset, we use sequences from embedding space from one digit to another obtained via Conditional Variational Autoencoder~\cite{sohn2015learning} (CVAE). Then we take two points corresponding to a certain pair of digits and also add the points from the line connecting two initial points. 
For each sequence, we apply a decoder to get sequence images that closely reflect a corresponding digit.
In the dataset, we put sequences with and without a change point. 
Our data consist of 1000 sequences with a length 64.
The dataset is balanced: the number of sequences with and without changes is equal. 

\subsection{Metrics}

We use metrics commonly used for the evaluation of change point detection algorithms~\cite{van2020evaluation}: $F_1$ score, Covering and Area under the detection curve (Area)~\cite{romanenkova2021principled}.
Some of them are inspired by Image segmentation, as these two problem statements share a lot in common~\cite{arbelaez2010contour}.

Better methods have bigger $F_1$ scores and Covering, but smaller Area. We refer an interested reader to the review~\cite{van2020evaluation} for a more detailed discussion of the metrics.

\begin{table}   \label{table:main}
    \begin{tabular}{cccc}
        \hline
            & ~$F_1$ score $\uparrow$ & ~Covering $\uparrow$& ~Area $\downarrow$ \\
        \hline
            RNN-based, CPD loss & ~$0.660 \pm 0.011 $ & ~ $0.973 \pm 0.003 $ & ~ $244 \pm 30 $ \\
            RNN-based, BCE loss & ~${0.659 \pm 0.016 }$ & ~ $0.974 \pm 0.003 $ & ~ $234 \pm 37 $ \\
            Transformer without mask, CPD loss & ~$0.544 \pm 0.074 $ & ~ $0.973 \pm 0.002 $ & ~ $\underline{217 \pm 9}$ \\
            Transformer without mask, BCE loss & ~$0.560 \pm 0.127 $ & ~ $0.972 \pm 0.005 $ & ~ $\underline{229 \pm 18}$ \\
            Low-triangular mask, CPD loss & ~$\underline{0.677 \pm {0.026}}$ & ~ $0.971 \pm 0.003 $ & ~ $240 \pm 20 $ \\
            Low-triangular mask, BCE loss & ~$\underline{0.664 \pm 0.016 }$ & ~ $0.973 \pm 0.004 $ & ~ $\underline{229 \pm 26 }$ \\
            2-diagonal mask, CPD loss & ~${0.668 \pm 0.021} $ & ~ ${0.976} \pm {0.002} $ & ~ $\textbf{211} \pm \textbf{19}$ \\
            2-diagonal mask, BCE loss & ~$0.656 \pm 0.022 $ & ~ $\underline{0.977 \pm 0.003} $ & ~ $\textbf{219} \pm \textbf{17} $ \\
            8-diagonal mask, CPD loss & ~${0.668 \pm 0.027 }$ & ~ $\textbf{0.978} \pm \textbf{0.004} $ & ~ ${221} \pm {30}$ \\
            1-diag. + 8-lower-triang., BCE loss & ~$0.656 \pm 0.022 $ & ~ $\textbf{0.978} \pm \textbf{0.003} $ & ~ $232 \pm 18 $ \\
            1-diag. + 8-lower-triang., CPD loss & ~$\textbf{0.679} \pm \textbf{0.024} $ & ~ $\underline{0.976 \pm 0.002} $ & ~ ${229 \pm 22 }$ \\
            3-diag. + 32-lower-triang., BCE loss & ~$\textbf{0.678} \pm \textbf{0.013} $ & ~ ${0.974} \pm {0.003} $ & ~ $232 \pm 28 $ \\
        \hline
    \end{tabular}
    \caption{Comparison of the performance of different approaches, mean $\pm$ std. The best value \textit{for each type of loss} are in \textbf{bold}. The second-best value for each type of loss are  \underline{underlined}.}
  
\end{table}

\subsection{Main results}
We compare methods based on the attention mechanism and transformers with more common approaches for sequential data processing. 
Among different attention mechanisms, we consider sequential attention with low triangular attention matrix, diagonal and tri-diagonal attention along.

We present our main results in Table~\ref{table:main}. As we see from the results, usage of specific loss with attention improves our results compared to Recurrent Neural Network and Binary Cross-Entropy loss. 

Along with the table, we present figures with a variation of the main hyperparameter of our approach: how many off-diagonal elements we use. 
Figures ~\ref{Fig:covering_bce} -- \ref{Fig:area_cpd} in \ref{Appendix} demonstrates the dynamic of the performance of our methods with respect to this hyperparameters.
We consider $F_1$ score, Area under the detection curve and Covering. 

All the results for both the table and figures were obtained by averaging over 10 runs. For all metrics, we see an improvement compared to a vanilla RNN approach. We also see that BCE loss is typically less stable than CPD loss. The diagonal mask with lower-triangular seems to perform the best among introduced approaches if the correct window size is used.

\section{Acknowledgements}

The work of Alexey Zaytsev was supported by the Russian Foundation for Basic Research grant  20-01-00203. 
The work of Evgenya Romanenkova was supported  by the Russian Science Foundation (project 20-71-10135). 

\section{Conclusions}

We propose a method of change-point detection based on an attention mechanism. We prove that choosing an attention matrix with respect to the nature of the task solve the change point detection problem more efficiently. Choosing any reasonable matrices with a combination of principled loss functions improves results in all metrics. However, the most successful methods are diagonal and diagonal with a small lower-triangular tail. We show that our approach outperforms RNN-based state-of-the-art on up to 15\% area under the detection curve, which means faster and accurate predictions. 

\printbibliography[ heading=bibintoc,]
\newpage
\section*{\\Appendix}\label{Appendix}
There are results from variation of the main hyper-parameter of our approach.

\begin{figure}
   \begin{minipage}{1\textwidth}
     \centering
     \def\svgwidth{\columnwidth}
        \includegraphics[width=11.5cm]{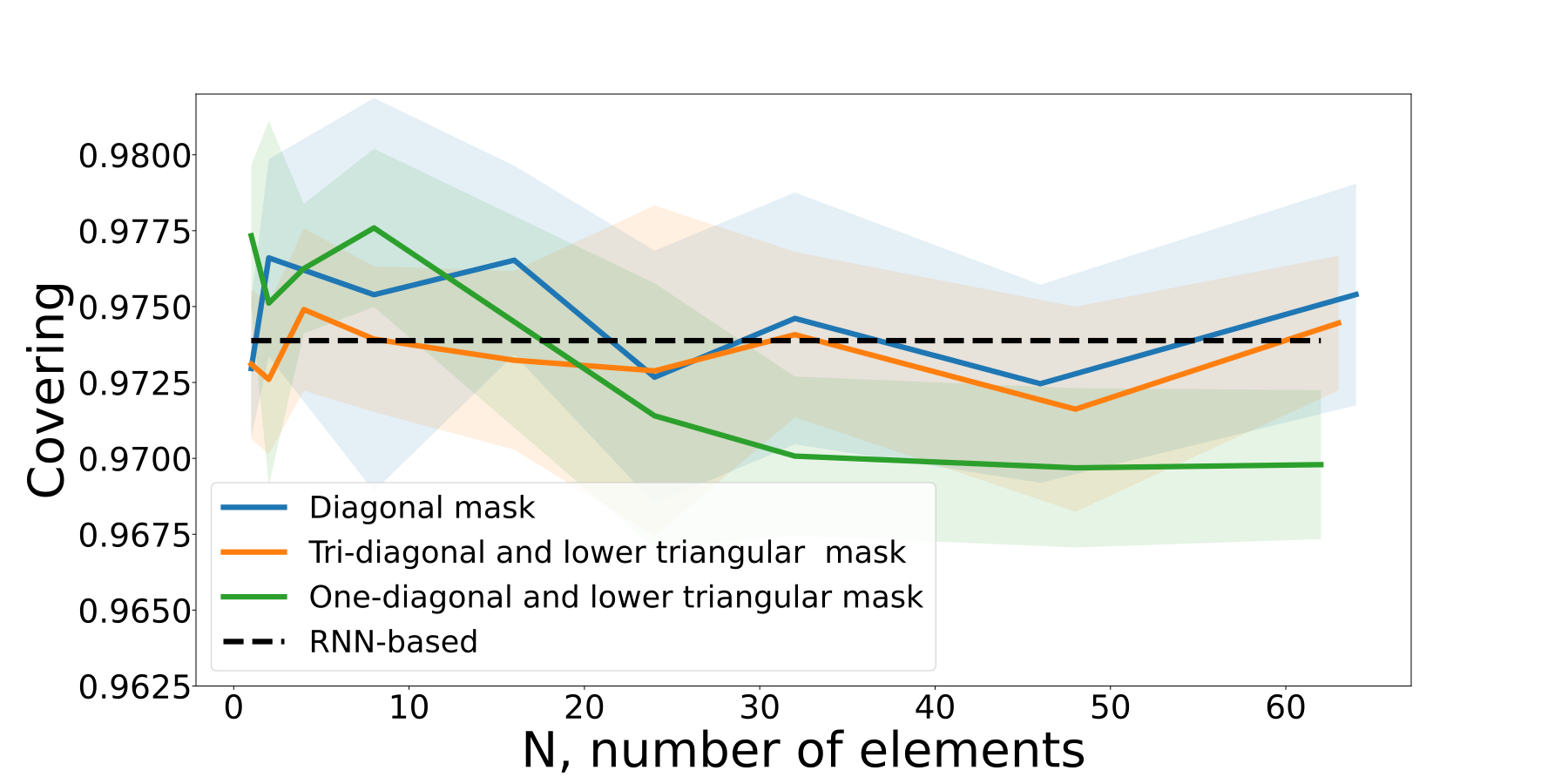}
     \caption{Dynamic of covering for different masks size, BCE loss}\label{Fig:covering_bce}
   \end{minipage}\hfill
   \begin{minipage}{1\textwidth}
     \centering
        \def\svgwidth{\columnwidth}
        \includegraphics[width=11.5cm]{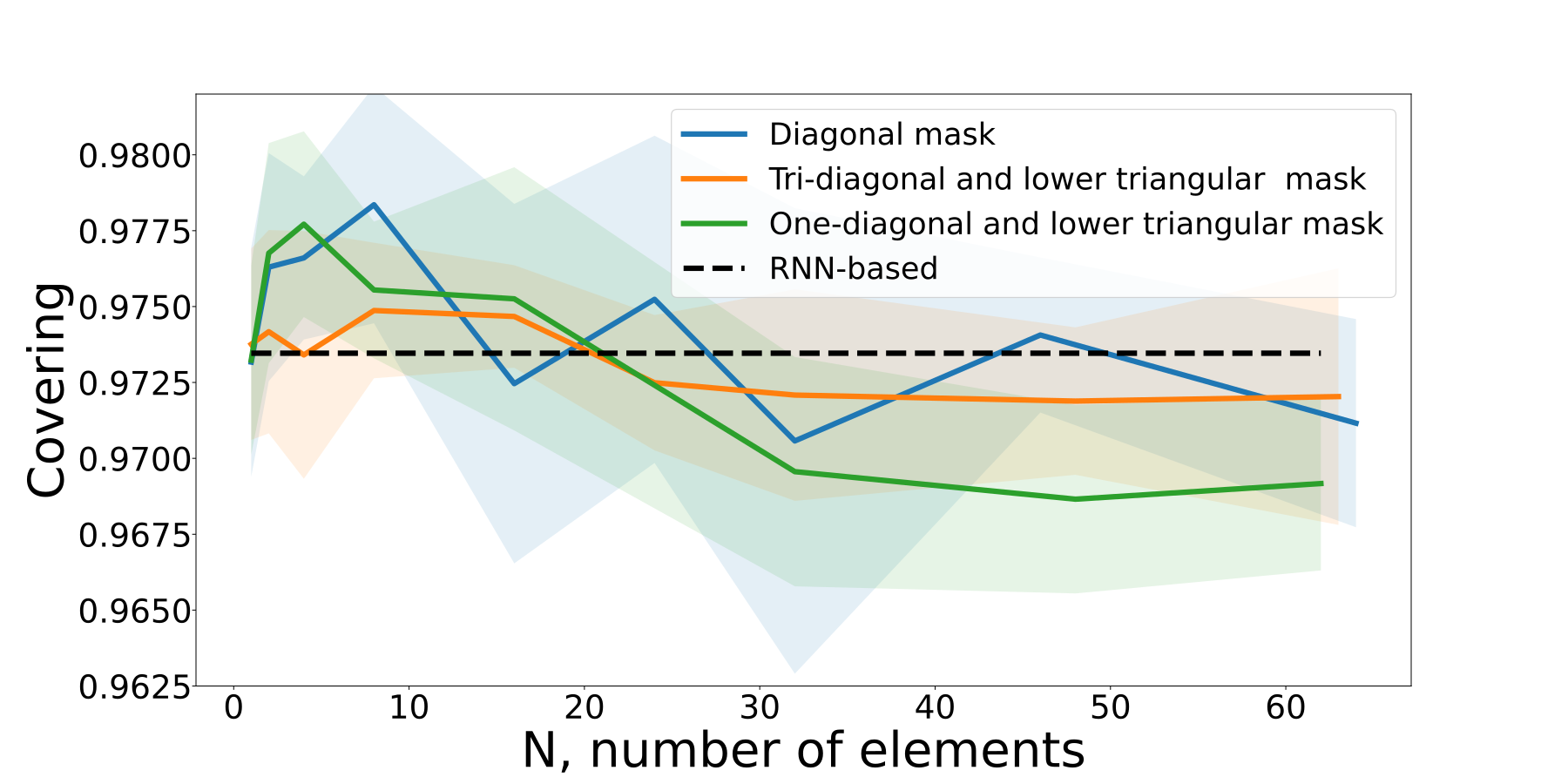}
     \caption{Dynamic of covering for different masks size, CPD loss}\label{Fig:covering_cpd}
   \end{minipage}
\end{figure}

\begin{figure}
   \begin{minipage}{1.\textwidth}
     \centering
     \def\svgwidth{\columnwidth}
         \includegraphics[width=11.5cm]{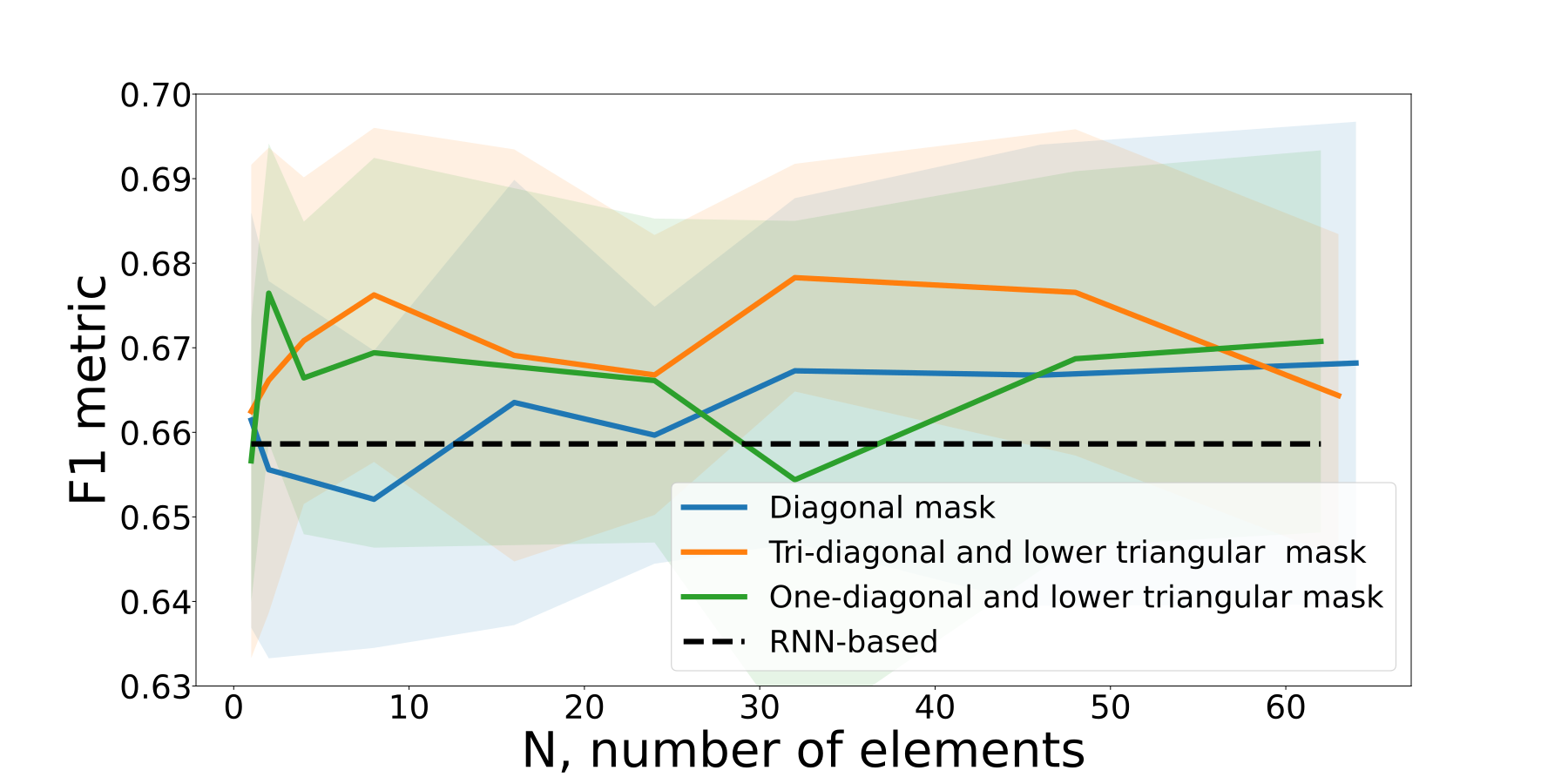}
     \caption{Dynamic of F1-score for different masks size, BCE loss}\label{Fig:f1metric_bce}
   \end{minipage}\hfill
   \begin{minipage}{1.\textwidth}
     \centering
        \def\svgwidth{\columnwidth}
        \includegraphics[width=11.5cm]{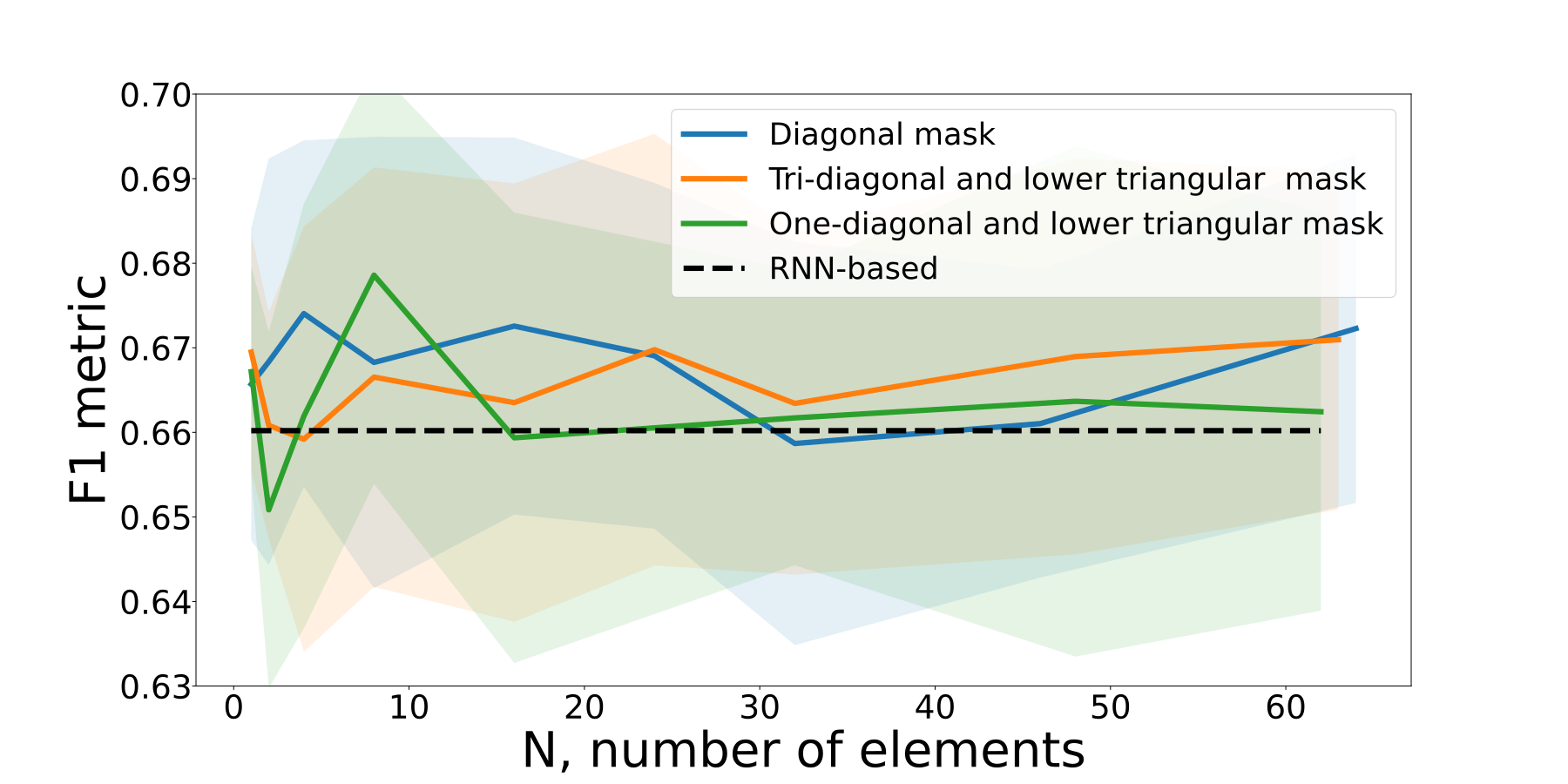}
     \caption{Dynamic of F1-score for different masks size, CPD loss}\label{Fig:f1metric_cpd}
   \end{minipage}
\end{figure}

\begin{figure}
   \begin{minipage}{1.\textwidth}
     \centering
     \def\svgwidth{\columnwidth}
        \includegraphics[width=11.5cm]{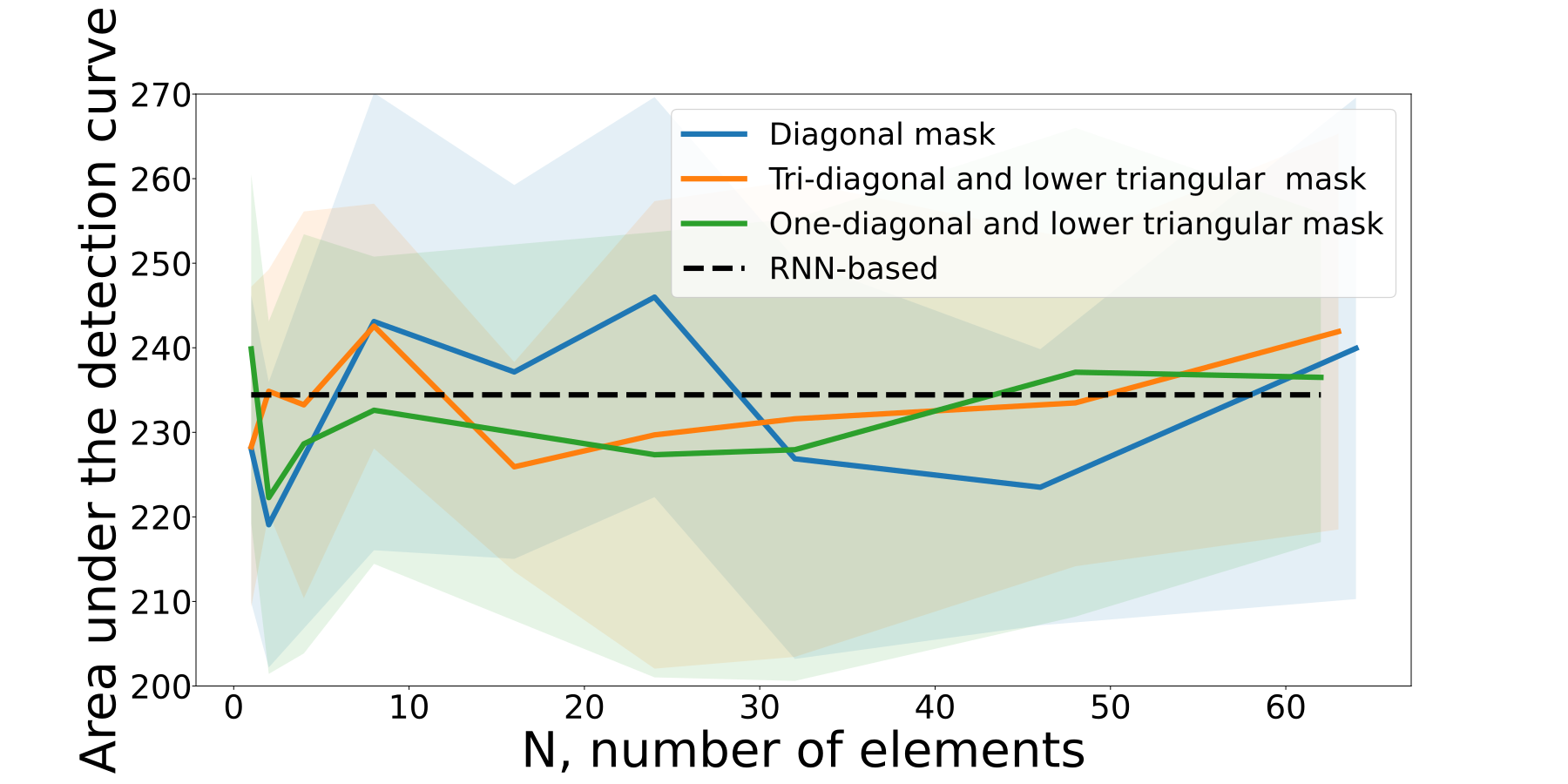}
     \caption{Dynamic of area under the detection curve for different masks, BCE loss}\label{Fig:area_bce}
   \end{minipage}\hfill
   \begin{minipage}{1.\textwidth}
     \centering
        \def\svgwidth{\columnwidth}
        \includegraphics[width=11.5cm]{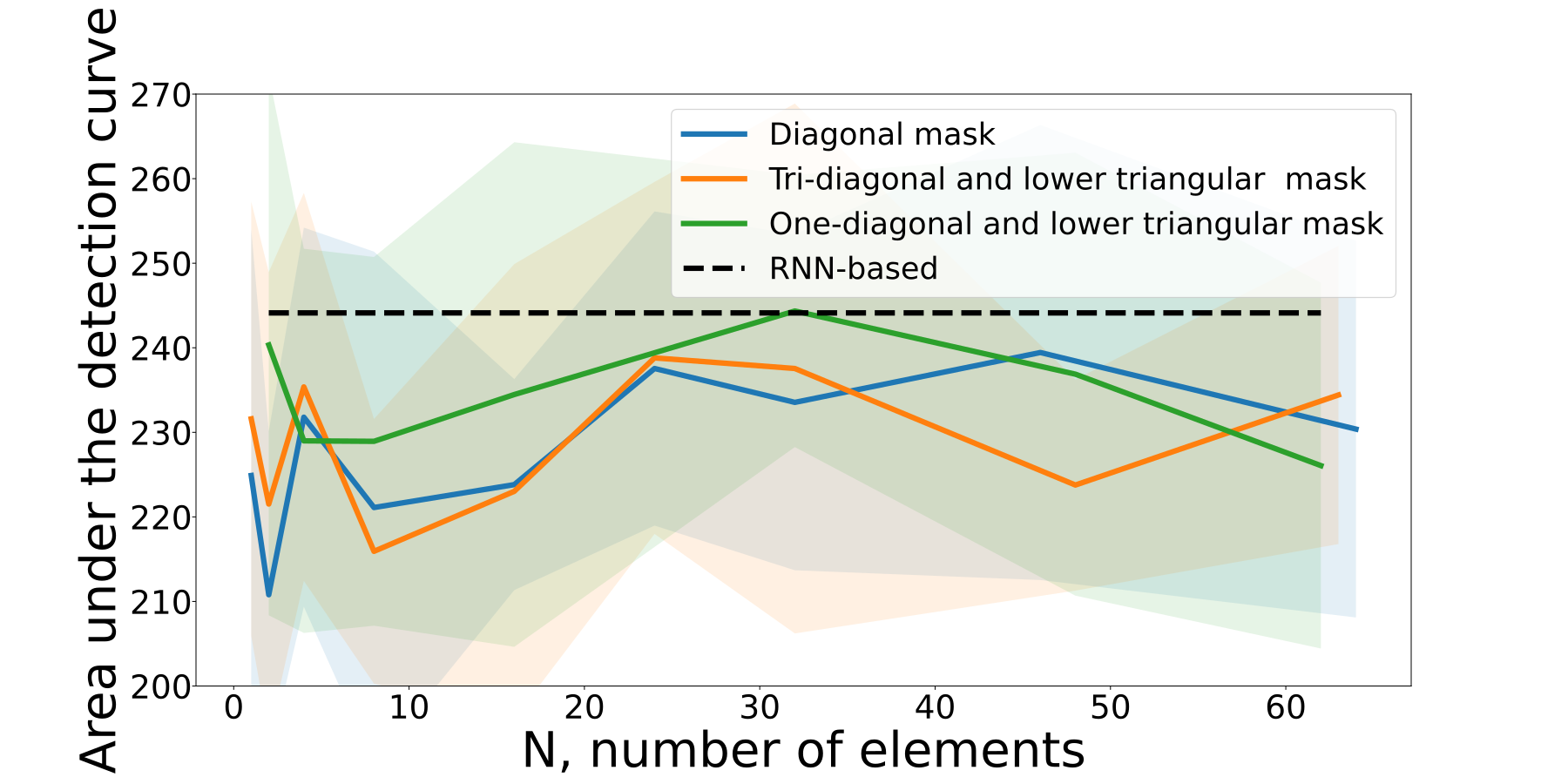}
     \caption{Dynamic of area under the detection curve for different masks, CPD loss}\label{Fig:area_cpd}
   \end{minipage}
\end{figure}

\end{document}